\documentclass[11pt,a4paper]{article}
\usepackage[hyperref]{naaclhlt2019}
\usepackage{times}
\usepackage{latexsym}

\usepackage{url}

\usepackage{graphicx}
\newtheorem{theorem}{Theorem}

\aclfinalcopy 

\title{Fixed-Size Ordinally Forgetting Encoding Based\\ Word Sense Disambiguation}

\author{Xi Zhu, Mingbin Xu, Hui Jiang \\
	Department of Electrical Engineering and Computer Science \\
	Lassonde School of Engineering, York University \\
	4700 Keele Street, Toronto, Ontario, Canada\\
	{\tt \{xzhu, xmb, hj\}@eecs.yorku.ca }
}

\date{}

\begin{document}
\maketitle
\begin{abstract}
  In this paper, we present our method of using fixed-size ordinally forgetting encoding (FOFE) to solve the word sense disambiguation (WSD) problem. FOFE enables us to encode variable-length sequence of words into a theoretically unique fixed-size representation that can be fed into a feed forward neural network (FFNN), while keeping the positional information between words. In our method, a FOFE-based FFNN is used to train a pseudo language model over unlabelled corpus, then the pre-trained language model is capable of abstracting the surrounding context of polyseme instances in labelled corpus into context embeddings. Next, we take advantage of these context embeddings towards WSD classification. We conducted experiments on several WSD data sets, which demonstrates that our proposed method can achieve comparable performance to that of the state-of-the-art approach at the expense of much lower computational cost.
\end{abstract}

\section{Introduction}

Words with multiple senses commonly exist in many languages. For example, the word \textit{bank} can either mean a ``financial establishment'' or ``the land alongside or sloping down to a river or lake'', based on different contexts. Such a word is called a ``polyseme''. The task to identify the meaning of a polyseme in its surrounding context is called word sense disambiguation (WSD). Word sense disambiguation is a long-standing problem in natural language processing (NLP), and has broad applications in other NLP problems such as machine translation \citep{Taghipour:15}. Lexical sample task and all-word task are the two main branches of WSD problem. The former focuses on only a pre-selected set of polysemes whereas the later intends to disambiguate every polyseme in the entire text. Numerous works have been devoted in WSD task, including supervised, unsupervised, semi-supervised and knowledge based learning \citep{Iacobacci:16}. Our work focuses on using supervised learning to solve all-word WSD problem.

    Most supervised approaches focus on extracting features from words in the context. Early approaches mostly depend on hand-crafted features. For example, IMS by \citet{Zhong:10} uses POS tags, surrounding words and collections of local words as features. These approaches are later improved by combining with word embedding features \citep{Taghipour:15}, which better represents the words' semantic information in a real-value space. However, these methods neglect the valuable positional information between the words in the sequence \citep{Kageback16}. The bi-directional Long-Short-Term-Memory (LSTM) approach by \citet{Kageback16} provides one way to leverage the order of words. Recently, \citet{Yuan:16} improved the performance by pre-training a LSTM language model with a large unlabelled corpus, and using this model to generate sense vectors for further WSD predictions. However, LSTM significantly increases the computational complexity during the training process.

    The development of the so called ``fixed-size ordinally forgetting encoding'' (FOFE) has enabled us to consider more efficient method. As firstly proposed in \citep{Zhang:15}, FOFE provides a way to encode the entire sequence of words of variable length into an almost unique fixed-size representation, while also retain the positional information for words in the sequence. FOFE has been applied to several NLP problems in the past, such as language model \citep{Zhang:15}, named entity recognition \citep{Xu:16}, and word embedding \citep{Sanu:16}. The promising results demonstrated by the FOFE approach in these areas inspired us to apply FOFE in solving the WSD problem. In this paper, we will first describe how FOFE is used to encode sequence of any length into a fixed-size representation. Next, we elaborate on how a pseudo language model is trained with the FOFE encoding from unlabelled data for the purpose of context abstraction, and how a classifier for each polyseme is built from context abstractions of its labelled training data. Lastly, we provide the experiment results of our method on several WSD data sets to justify the equivalent performance as the state-of-the-art approach.

\section{Fixed-size Ordinally Forgetting Encoding}

    The fact that human languages consist of variable-length sequence of words requires NLP models to be able to consume variable-length data. RNN/LSTM addresses this issue by recurrent connections, but such recurrence consequently increases the computational complexity. On the contrary, feed forward neural network (FFNN) has been widely adopted in many artificial intelligence problems due to its powerful modelling ability and fast computation, but is also limited by its requirement of fixed-size input. FOFE aims at encoding variable-length sequence of words into a fixed-size representation, which subsequently can be fed into an FFNN. 
    
    Given vocabulary $V$ of size $|V|$, each word can be represented by a one-hot vector. FOFE can encode a sequence of words of any length using linear combination, with a forget factor to reflect the positional information. For a sequence of words $S=w_1, w_2, .., w_T$ from V, let $\textbf{e}_{i}$ denote the one-hot representation for the $i^{th}$ word, then the FOFE code of S can be recursively obtained using following equation (set $\textbf{z}_{0} = \textbf{0}$):
    $$\textbf{z}_{t}= \alpha \cdot \textbf{z}_{t-1} + \textbf{e}_{t} \quad (1 \le t \le T)$$
    where $\alpha$ is a constant between 0 and 1, called forgetting factor. For example, assuming A, B, C are three words with one-hot vectors $[1,0,0]$, $[0,1,0]$, $[0,0,1]$ respectively. The FOFE encoding from left to right for {ABC} is [$\alpha^2$,$\alpha$,1] and for {ABCBC} is [$\alpha$,$\alpha+\alpha$,$1+\alpha$]. It becomes evident that the FOFE code is in fixed size, which is equal to the size of the one-hot vector, regardless of the length of the sequence $S$.
    
    The FOFE encoding has the property that the original sequence can be unequivocally recovered from the FOFE encoding. According to \citet{Zhang:15}, the uniqueness for the FOFE encoding of a sequence is confirmed by the following two theorems: 
    
    \begin{theorem}
        If the forgetting factor $\alpha$ satisfies $0 \leq \alpha < 0.5$, FOFE is unique for any sequence of finite length $T$ and any countable vocabulary $V$.
    \end{theorem}
    
    \begin{theorem} \label{theo-2}
        If the forgetting factor $\alpha$ satisfies $0.5 \leq \alpha \leq 1$, FOFE is almost unique for any finite value of $T$ and vocabulary $V$, except only a finite set of countable choices of $\alpha$. 
    \end{theorem}
        
    
    
    Even for situations described by Theorem \ref{theo-2} where uniqueness is not strictly guaranteed, the probability for collision is extremely low in practice. Therefore, FOFE can be safely considered as an encoding mechanism that converts variable-length sequence into a fixed-size representation theoretically without any loss of information.

\section{Methodology}
\begin{figure}[h]
        \centering
        \includegraphics[width=0.5\textwidth]{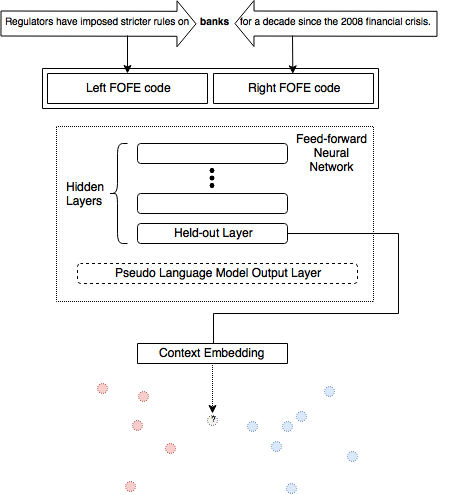}
        \caption{Context abstraction through FOFE-based pseudo language model and WSD classification over context embeddings}
        \label{fig1}
    \end{figure}

    The linguistic distribution hypothesis states that words that occur in close contexts should have similar meaning \citep{Harris:1954}. It implies that the particular sense of a polyseme is highly related to its surrounding context. Moreover, human decides the sense of a polyseme by firstly understanding its occurring context. Likewise, our proposed model has two stages, as shown in Figure~\ref{fig1}: training a FOFE-based pseudo language model that abstracts context as embeddings, and performing WSD classification over context embeddings.
    
\subsection{FOFE-based Pseudo Language Model}
    
    A language model is trained with large unlabelled corpus by \citet{Yuan:16} in order to overcome the shortage of WSD training data. A language model represents the probability distribution of a given sequence of words, and it is commonly used in predicting the subsequent word given preceding sequence. \citet{Zhang:15} proposed a FOFE-based neural network language model by feeding FOFE code of preceding sequence into FFNN. WSD is different from language model in terms of that the sense prediction of a target word depends on its surrounding sequence rather than only preceding sequence. Hence, we build a pseudo language model that uses both preceding and succeeding sequence to accommodate the purpose of WSD tasks.
        
    The preceding and succeeding sequences are separately converted into FOFE codes. As shown in Figure~\ref{fig1}, the words preceding the target word are encoded from left to right as the left FOFE code, and the words succeeding the target word are encoded from right to left as the right FOFE code. The forgetting factor that underlies the encoding direction reflects the reducing relevance of a word due to the increasing distance relative to the target word. Furthermore, the FOFE is scalable to higher orders by merging tailing partial FOFE codes. For example, a second order FOFE of sequence $S=w_1, w_2, .., w_T$ can be obtained as $[z_{T-1}, z_T]$. Lastly, the left and right FOFE codes are concatenated into one single fixed-size vector, which can be fed into an FFNN as an input.
    
    FFNN is constructed in fully-connected layers. Each layer receives values from previous layer as input, and produces values through a function over weighted input values as its output. FFNN increasingly abstracts the features of the data through the layers. As the pseudo language model is trained to predict the target word, the output layer is irrelevant to WSD task and hence can be discarded. However, the remaining layers still have learned the ability to generalize features from word to context during the training process. The values of the held-out layer (the second last layer) are extracted as context embedding, which provides a nice numerical abstraction of the surrounding context of a target word.
    
\subsection{WSD Classification}
    
    Words with the same sense mostly appear in similar contexts, hence the context embeddings of their contexts are supposed to be close in the embedding space. As the FOFE-based pseudo language model is capable of abstracting surrounding context for any target word as context embeddings, applying the language model on instances in annotated corpus produces context embeddings for senses. 
    
    A classifier can be built for each polyseme over the context embeddings of all its occurring contexts in the training corpus. When predict the sense of a polyseme, we similarly extract the context embedding from the context surrounding the predicting polyseme, and send it to the polyseme's classifier to decide the sense. If a classifier cannot be built for the predicting polyseme due to the lack of training instance, the first sense from the dictionary is used instead.
    
    For example, word $w$ has two senses $s_i$ for $i=1,2$ occurring in the training corpus, and each sense has $n_i$ instances. The pseudo language model converts all the instances into context embeddings $\textbf{c}^i_j$ for $j=1,\ldots,n_i$, and these embeddings are used as training data to build a classifier for $w$. The classifier can then be used to predict the sense of an instance of $w$ by taking the predicting context embedding $\textbf{c}'$.
    
    The context embeddings should fit most traditional classifiers, and the choice of classifier is empirical. \citet{Yuan:16} takes the average over context embeddings to construct sense embeddings $\textbf{s}_i = \frac{\sum_{j=i}{\textbf{c}^i_j}}{n_i}$, and selects the sense whose sense embedding is closest to the predicting context embedding measured by cosine similarity. In practice, we found k-nearest neighbor (kNN) algorithm, which predicts the sense to be the majority of k nearest neighbors, produces better performance on the context embeddings produced by our FOFE-based pseudo language model.
    
    
\begin{table*}
\begin{center}
\resizebox{\textwidth}{!}{
    \begin{tabular}{|l|rr|r|rr|}
    \hline 
    \bf Model & \bf Corpus Size & \bf Vocab. & \bf Training Time & \bf Senseval2 & \bf SemEval13  \\ 
    \hline
    IMS $^\ast$ & - & - & - & 0.625 & - \\
    IMS + Word2vec $^\ast$ & - & - & - & 0.634 & -  \\
    \hline 
    LSTM \citep{Yuan:16} & 100B & 1M & - & 0.736 & 0.670   \\
    LSTM \citep{Le:16} & 2B & 1M & 4.5 months & 0.700 & 0.666   \\
    \hline 
    LSTM (our training) $^\dagger$ & 0.8B & 100K & 2 weeks & 0.661 & 0.633 \\
    \textbf{FOFE (this work)} & 0.8B & 100K & 3 days & 0.693 & 0.650  \\
    \hline
    \end{tabular}
}
\end{center}
    \caption{\label{senseval-result} The corpus size,  vocabulary size and training time when pre-training the language models, and F1 scores of different models on multiple WSD tasks using SemCor as training data. The asterisk ($\ast$) indicates the results are from \citep{Iacobacci:16}. Our training ($\dagger$) uses code published by \citep{Le:16} with Google1B \citep{Google1B} as training data.}
\end{table*}

\section{Experiment}
    To evaluate the performance of our proposed model, we implemented our model using Tensorflow \citep{tensorflow} and conducted experiments on standard SemEval data that are labelled by senses from WordNet 3.0 \citep{wordnet}. We built the classifier using SemCor \citep{semcor} as training corpus, and evaluated on Senseval2 \citep{senseval2}, and SemEval-2013 Task 12 \citep{semeval2013}.

\subsection{Experiment settings}
    When training our FOFE-based pseudo language model, we use Google1B \citep{Google1B} corpus as the training data, which consists of approximately 0.8 billion words. The 100,000 most frequent words in the corpus are chosen as the vocabulary. The dimension of word embedding is chosen to be 512. During the experiment, the best results are produced by the 3rd order pseudo language model. The concatenation of the left and right 3rd order FOFE codes leads to a dimension of 512 * 3 * 2 = 3072 for the FFNN's input layer. Then we append three hidden layers of dimension 4096. Additionally, we choose a constant forgetting factor $\alpha = 0.7$ for the FOFE encoding and $k=8$ for our k-nearest neighbor classifier.

\subsection{Results}
    Table~\ref{senseval-result} presents the micro F1 scores from different models. Note that we use a corpus with 0.8 billion words and vocabulary of 100,000 words when training the language model, comparing with \citet{Yuan:16} using 100 billion words and vocabulary of 1,000,000 words. The context abstraction using the language model is the most crucial step. The sizes of the training corpus and vocabulary significantly affect the performance of this process, and consequently the final WSD results. However, \citet{Yuan:16} did not publish the 100 billion words corpus used for training their LSTM language model.

    Recently, \citet{Le:16} reimplemented the LSTM-based WSD classifier. The authors trained the language model with a smaller corpus Gigaword \citep{gigaword} of 2 billion words and vocabulary of 1 million words, and reported the performance.
    Their published code also enabled us to train an LSTM model with the same data used in training our FOFE model, and compare the performances at the equivalent conditions.

    Additionally, the bottleneck of the LSTM approach is the training speed. The training process of the LSTM model by \citet{Le:16} took approximately 4.5 months even after applying optimization of trimming sentences, while the training process of our FOFE-based model took around 3 days to produce the claimed results.

\section{Conclusion}

    In this paper, we propose a new method for word sense disambiguation problem, which adopts the fixed-size ordinally forgetting encoding (FOFE) to convert variable-length context into almost unique fixed-size representation. A feed forward neural network pseudo language model is trained with FOFE codes of large unlabelled corpus, and used for abstracting the context embeddings of annotated instance to build a k-nearest neighbor classifier for every polyseme. Compared to the high computational cost induced by LSTM model, the fixed-size encoding by FOFE enables the usage of a simple feed forward neural network, which is not only much more efficient but also equivalently promising in numerical performance.
    
\bibliography{naaclhlt2019}
\bibliographystyle{acl_natbib}
\end{document}